\documentclass{article}
\usepackage{spconf,amsmath,graphicx, epsfig, array, subfigure, float}
\usepackage{caption}
\usepackage{amssymb}
\usepackage{tabu,multirow}
\usepackage{booktabs}

\makeatletter
\newcommand{\thickhline}{%
    \noalign {\ifnum 0=`}\fi \hrule height 1pt
    \futurelet \reserved@a \@xhline
}
\usepackage{arydshln}

\def\L{{\cal L}}

\newcommand\norm[1]{\left\lVert#1\right\rVert}
\newcommand\blfootnote[1]{%
  \begingroup
  \renewcommand\thefootnote{}\footnote{#1}%
  \addtocounter{footnote}{-1}%
  \endgroup
}

\title{ORTHOGONALITY CONSTRAINED MULTI-HEAD ATTENTION \\ FOR KEYWORD SPOTTING}
\name{Mingu Lee, Jinkyu Lee, Hye Jin Jang, Byeonggeun Kim, Wonil Chang and Kyuwoong Hwang}
\address{Qualcomm AI Research, Qualcomm Korea YH}

\begin{document}
%
\maketitle
\begin{abstract}
Multi-head attention mechanism is capable of learning various representations from sequential data while paying attention to different subsequences, e.g., word-pieces or syllables in a spoken word. From the subsequences, it retrieves richer information than a single-head attention which only summarizes the whole sequence into one context vector. However, a naive use of the multi-head attention does not guarantee such richness as the attention heads may have positional and representational redundancy. In this paper, we propose a regularization technique for multi-head attention mechanism in an end-to-end neural keyword spotting system. Augmenting regularization terms which penalize positional and contextual non-orthogonality between the attention heads encourages to output different representations from separate subsequences, which in turn enables leveraging structured information without explicit sequence models such as hidden Markov models. In addition, intra-head contextual non-orthogonality regularization encourages each attention head to have similar representations across keyword examples, which helps classification by reducing feature variability. The experimental results demonstrate that the proposed regularization technique significantly improves the keyword spotting performance for the keyword ``Hey Snapdragon".
\end{abstract}


\blfootnote{Qualcomm AI Research is an initiative of Qualcomm Technologies, Inc.}

\section{Introduction}
\label{sec:intro}



Keyword spotting has recently been an essential function of consumer devices, such as mobile phones and smart speakers, because it provides a natural way of voice user interface. It is mainly used for detecting pre-defined keywords, e.g., ``Alexa", ``Hey Siri", and ``OK Google" for getting devices ready to process users' following commands or queries. Despite the widespread use of this technology in various devices today, it is still a challenging problem due to requiring low false rejection rate (FRR) and false alarm rate (FAR) while operating with small memory footprint and low power consumption.

In the previous studies, keyword/filler hidden Markov models (HMMs) were proposed, which explicitly model the acoustic characteristics of non-keyword general speech (filler) as well as the target keyword speech \cite{rohlicek1989continuous}. With the recent advances in deep learning, Gaussian mixture models in the HMMs were replaced with various neural network architectures, such as feed forward deep neural networks, convolutional neural networks, and convolutional recurrent neural networks (CRNNs) \cite{chen2014small, sainath2015convolutional, arik2017convolutional, lengerich2016end}. Although those deep learning-based approaches significantly improve the system performance by increasing modeling capacity, it still requires well predicted time-aligned labels.

Recently, a number of attention-based keyword spotting models have been proposed \cite{he2017streaming, shan2018attention, de2018neural}. While \cite{he2017streaming} used attention in an assistive form for biasing RNN-based decoders toward a keyword of interest, \cite{shan2018attention} aggressively deploy the attention mechanism proposed in \cite{chowdhury2017attention} for direct keyword feature representation with which a binary classifier discriminates keywords from nonkeywords. Since these approaches are based on basic single-head attention mechanism, it is natural to extend to use multi-head attention.
Multi-head attention \cite{vaswani2017attention, chiu2018state} is introduced for joint representation of information in different subspaces while attending to different positions of a sequence. However, as there is no explicit mechanism which guarantee such diversity either in positions and in representational subspaces, each attention head may contain redundant information which results in inefficiency of the network. \cite{li2018multi} proposed the three types of disagreement regularizations, i.e., disagreements on subspaces, attended positions and outputs, to explicitly encourage the diversity among attention heads based on the cosine similarity.

 

In this paper, we investigate the use of multi-head attention in keyword spotting tasks and propose an orthogonality constrained multi-head attention mechanism. The regularization is derived from the constraints of context and score vectors between attention heads such that they are orthogonal to each other, respectively. The regularization by inter-head orthogonality of context vectors and score vectors lets the attention heads have less redundancy to each other, while the regularization by intra-head non-orthogonality of context vectors lets them have consistency across samples for the given task. Regularization presented in this work is related to \cite{li2018multi} while it is more oriented to speech data and keyword spotting tasks. We show that the proposed regularization techniques improve the keyword detection performance by reducing the false rejection rates with only a small amount of increase in the model size.

\begin{figure}[!t]
  \centering
  \includegraphics[width=8.9cm]{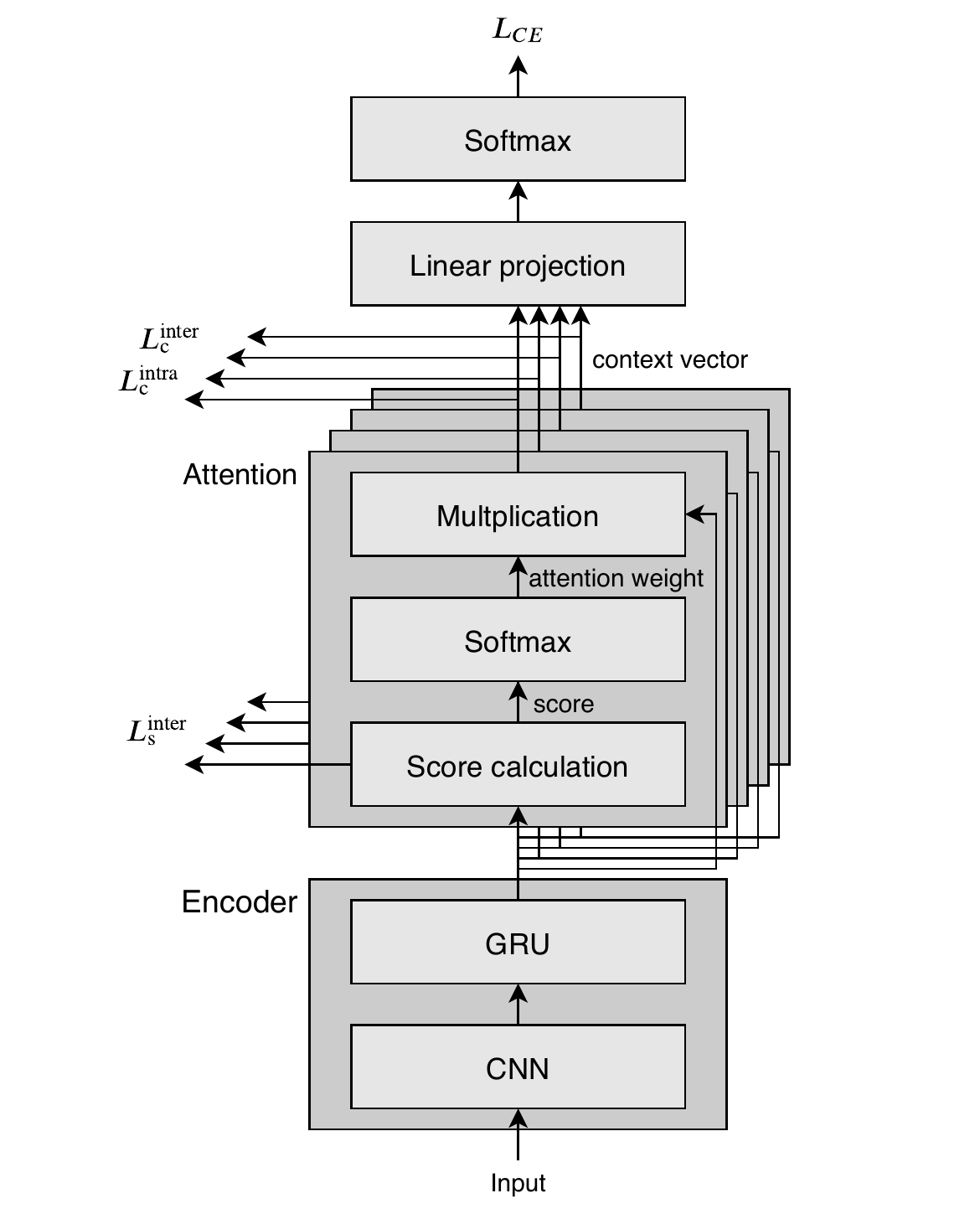}
  \vspace*{-10pt}
  \caption{The end-to-end keyword spotting network.}
  \vspace*{-5pt}
  \label{fig:base_model_structure}
\end{figure}

\section{MULTI-HEAD ATTENTION-BASED END-TO-END MODEL FOR KEYWORD SPOTTING}
\label{sec:attention}

\subsection{Keyword spotting system description}
\label{subsec:base_system}

We extend the single-head attention-based end-to-end network structure for keyword spotting presented in \cite{shan2018attention} to multi-head attention based network as depicted in Fig. \ref{fig:base_model_structure}. The encoder takes an acoustic feature $\mathbf{x}[t]$, $t=1,2,...,T$, the 40-dimensional Mel-filter bank energies extracted from 16 kHz sampled audio signals with per-channel energy normalization~\cite{wang2017trainable} where $t$ is the time frame index, as input and converts it into a hidden representation $\mathbf{h}[t]$. The encoder network consists of a canonical CRNN structure with convolutional and recurrent layers in sequence to capture spectral and temporal characteristics of the acoustic features. As a base model, we use one convolutional layer with a kernel size of $5\times20$ and a stride of $2\times1$ and one gated recurrent unit (GRU) layer with 64 hidden units as proposed in \cite{shan2018attention}. The encoder output vector $\mathbf{h}[t]$ is then processed by an attention mechanism in each attention head to produce a context vector $\mathbf{c}_i$ where $i$ denotes the attention head index. Using $H$ attention heads, the context vectors are concatenated as $\mathbf{c}=\big[ \mathbf{c}_1^\text{T}, \mathbf{c}_2^\text{T}, ..., \mathbf{c}_H^\text{T} \big]^\text{T}$ where $^\text{T}$ denotes the matrix transpose. Finally, the model performs binary classification with a linear transformation and a softmax operation on $\mathbf{c}$ to compute a posterior probability of a keyword state $y$ given input observation $\mathbf{x}$, $p(y|\mathbf{x})$. In the inference stage, we decide that a keyword is detected when the confidence $p(y|\mathbf{x})$ is larger than a pre-set threshold. Note that this system does not require any graph searching or frame-level alignment of training data, which largely simplifies both training and inference.

\subsection{Base attention mechanism}
\label{subsec:base_attention}

In each attention head, we use the nonlinear soft attention mechanism proposed in \cite{chowdhury2017attention} for speaker verification and adopted in \cite{shan2018attention} for keyword spotting. The attention weight $\alpha_i[t]$ for the $i$-th attention head at the $t$-th time frame is calculated~by
\begin{equation}
    \alpha_i[t] = \frac{\exp({e_i[t]})}{\sum_{\tau=1}^{T} \exp(e_i[\tau])},
\end{equation}
where the scalar score $e_i[t]$ is calculated by a nonlinear scoring function with the parameters shared across time
\begin{equation}
\label{eq:scoring_function}  e_i[t] = \mathbf{v}_i^{\text{T}}\tanh{\big(\mathbf{W}_{i}\mathbf{h}[t] + \mathbf{b}_i\big)}.
\end{equation}
The context vector $\mathbf{c}_i$, the output of each attention head, is then calculated by the weighted sum as follows:
\begin{equation}
    \mathbf{c}_i = \sum_{t=1}^{T}\alpha_i[t] \mathbf{h}[t].
\end{equation}

\section{ORTHOGONALITY REGULARIZED MULTI-HEAD ATTENTION}
\label{sec:regularized_multi_head_attention}

In speech recognition tasks including keyword spotting, although end-to-end neural networks are very attractive due to simplicity of their structures and learning procedures, hybrid systems often have competitive or even better performances as they use explicit sequence models for better leveraging structured information coming from speech subsequences, i.e., phonemes, syllables, or word-pieces \cite{luscher2019rwth}. In this perspective, the multi-head attention mechanism is considered as a promising alternative to capture the structured information from speech subsequences while keeping the end-to-end nature \cite{dong2019self,dong2018speech}.

\begin{figure}[!t]
    \begin{minipage}[b]{1.0\linewidth}
    \centering  
    \subfigure[]{\includegraphics[width=0.99\linewidth]{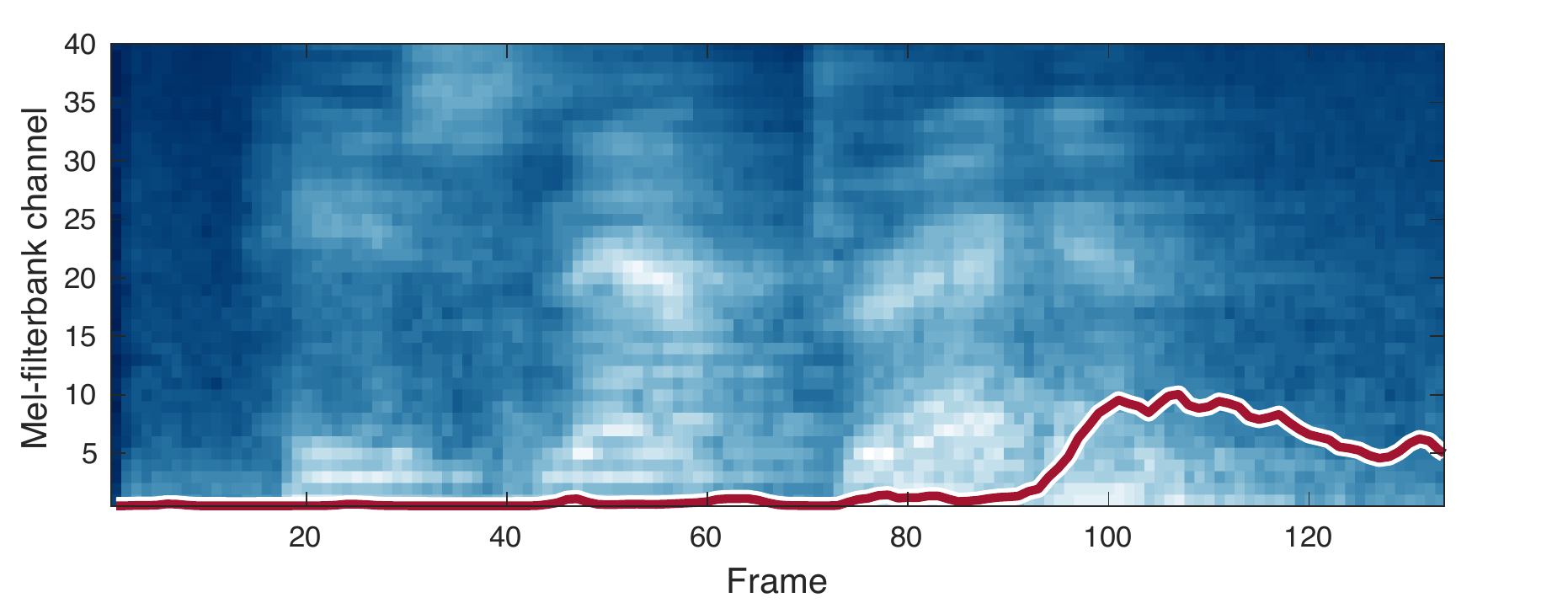}}
    \subfigure[]{\includegraphics[width=0.99\linewidth]{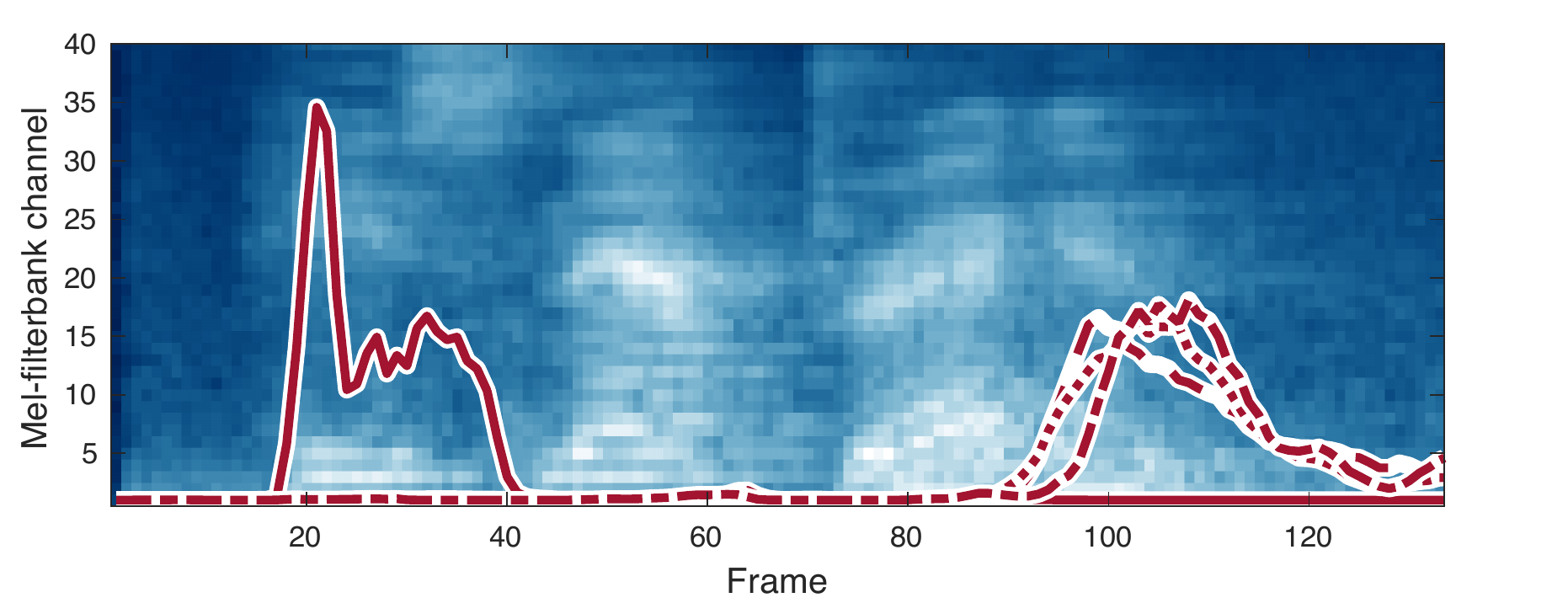}}
    \subfigure[]{\includegraphics[width=0.99\linewidth]{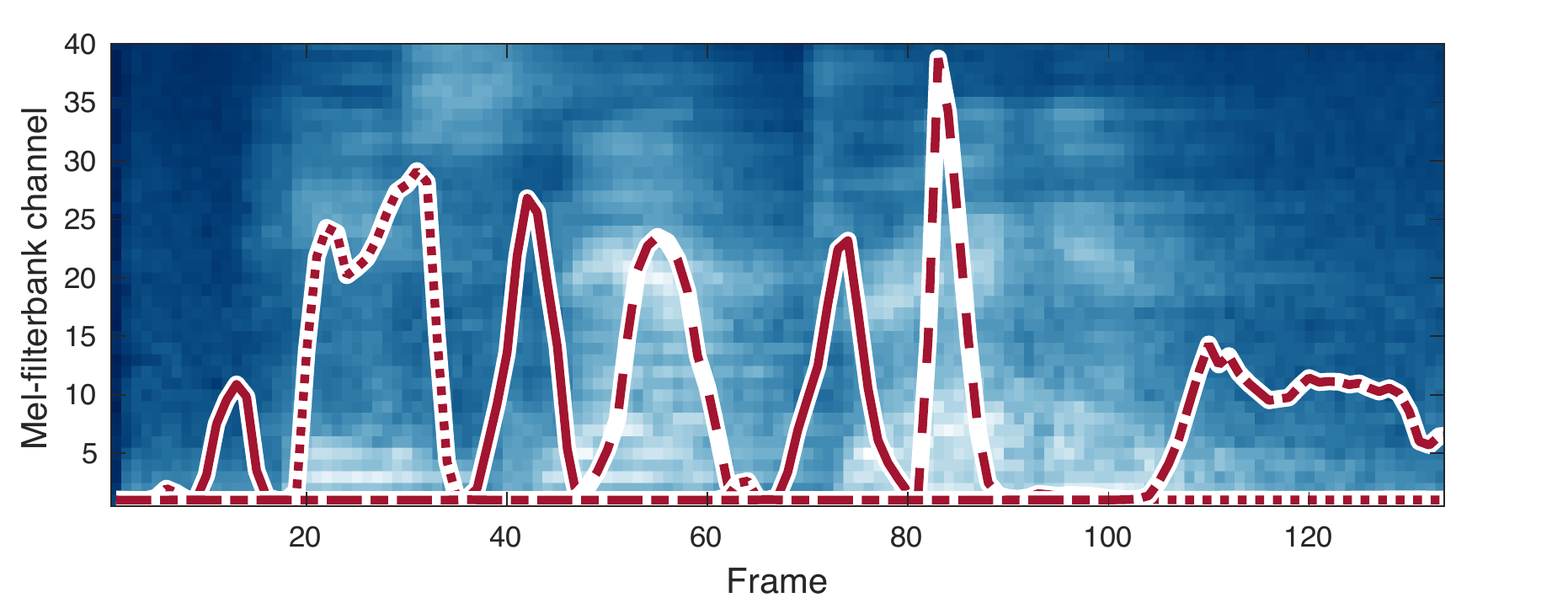}}
    \end{minipage}
    \caption{Attention weights overlaid on Mel-spectrogram of a ``Hey Snapdragon" utterance with the configurations of (a) single head attention, (b) 4-head attention without regularization, and (c) 4-head attention with the proposed regularization where all $\lambda$ values are set to 0.1. The attention weights are scaled with a factor of 320 for clear visibility.}
    \label{fig:attention_example}
\end{figure}

Multi-head attention, proposed in \cite{vaswani2017attention}, is capable of diverse learning of representations since different heads can pay attention to different positions in a sequence and give different representations. However, the diversity is not guaranteed by its natural form as they may have redundancy either in position and representation. Fig. \ref{fig:attention_example}(b) shows an example of multi-head attention weight distributions where 3 of them severely overlap to each other. For encouraging the diversity of the multi-head attention, \cite{li2018multi} proposed three types of disagreement regularization in the context of machine translation, i.e., disagreements on subspaces, attended positions and outputs, based on maximization of the negative cosine similarities. In this section, we propose a regularization technique for training the multi-head attention-based keyword spotting model by orthogonality constraints between attention heads.

\subsection{Inter-head orthogonality regularization}
\label{subsec:inter_head_orthogonality}

We argue that, to capture the temporally structured information in a sp eech input sequence, the attention heads should pay attention to different parts of the sequence and produce context outputs with minimal redundancy with each other. To achieve this, we introduce regularization of the multi-head attention by orthogonality constraints on context and score vectors between the attention heads. The problem is to find the network parameters that minimize the cross entropy loss $\L_{\text{CE}}$ subject to the orthogonality constraints $\mathbf{c}_i\perp\mathbf{c}_j$ and $\mathbf{e}_i\perp\mathbf{e}_j$ for each pair of $i\not=j$.
Suppose that we have a training batch of $N$ samples, then we define the regularization terms $\L_{\text{c}}^{\text{inter}}$ and $\L_{\text{s}}^{\text{inter}}$ by
\begin{equation}
\label{eq:inter_context}
    \L_{\text{c}}^{\text{inter}} = \frac{1}{N}\sum_{n=1}^{N}{\frac{1}{H(H-1)} \norm{\mathbf{C}^{(n)\text{T}}\mathbf{C}^{(n)} - \mathbf{I}_H}^2_\text{F}}
\end{equation}
\begin{equation}
\label{eq:inter_score}
    \L_{\text{s}}^{\text{inter}} = \frac{1}{N}\sum_{n=1}^{N}{\frac{1}{H(H-1)} \norm{\mathbf{E}^{(n)\text{T}}\mathbf{E}^{(n)} - \mathbf{I}_H}^2_\text{F}},
\end{equation}
where $n$ is the sample index, $H$ is the number of attention heads, $\norm{\cdot}^2_\text{F}$ denotes the Frobenius norm, and
\begin{equation}
    \mathbf{C}^{(n)}=\big[\overline{\mathbf{c}}_1^{(n)}, \overline{\mathbf{c}}_2^{(n)}, \dots \overline{\mathbf{c}}_H^{(n)}\big] \text{ with } \overline{\mathbf{c}}_i^{(n)}=\mathbf{c}_i^{(n)}/\norm{\mathbf{c}_i^{(n)}}
\end{equation}
\begin{equation}
    \mathbf{E}^{(n)}=\big[\overline{\mathbf{e}}_1^{(n)}, \overline{\mathbf{e}}_2^{(n)}, \dots \overline{\mathbf{c}}_H^{(n)}\big] \text{ with } \overline{\mathbf{e}}_i^{(n)}=\mathbf{e}_i^{(n)}/\norm{\mathbf{e}_i^{(n)}},
\end{equation}
are the context matrix and the score matrix, respectively, which consist of the normalized context vectors $\overline{\mathbf{c}}_i$ and the normalized score vectors $\overline{\mathbf{e}}_i$. 

One main difference from the output disagreement regularization in \cite{li2018multi} is that our system does not use value projection and thus directly compute the context vector from the encoder output $\mathbf{h}$ by multiplying the attention weights. Since the inter-head context orthogonality constraint can easily be satisfied by an orthogonal value projection in each head, regardless of the encoder outputs, we desire such orthogonality is achieved by the encoder network, not by the subspace projection. This encourages the encoder network to discriminateively represent different subsequences of a keyword utterance which results in better keyword detection.

\subsection{Intra-head non-orthogonality regularization}
\label{subsec:intra_head_nonorthogonality}

On the contrary, since each attention head finds a specific subsequence with similar content, the context vectors from the same attention head are expected to be similar across different samples. Thus, we augment a regularization term which maximizes the similarity or non-orthogonality of the context vectors between different samples from the same attention head as follows:
\begin{equation}
\label{eq:intra_context}
    \L_{\text{c}}^{\text{intra}} = \frac{1}{H}\sum_{i=1}^{H}{\frac{1}{N(N-1)} \norm{\widetilde{\mathbf{C}}_i^\text{T}\widetilde{\mathbf{C}}_i - \mathbf{I}_N}^2_\text{F}},
\end{equation}
where
\begin{equation}
    \widetilde{\mathbf{C}}_i=\big[\overline{\mathbf{c}}_i^{(1)}, \overline{\mathbf{c}}_i^{(2)}, \dots, \overline{\mathbf{c}}_i^{(N)}\big].
\end{equation}

Similar regularization to score vectors is not considered as the position of a subsequence attended by each attention head can vary from sample to sample.

\subsection{Selective regularization}
\label{subsec:selective_regularization}
Since the discussion about orthogonality and non-orthogonality constraints are only valid for positive data, i.e., keyword utterances, we modify (\ref{eq:inter_context}), (\ref{eq:inter_score}) and (\ref{eq:intra_context}) to be selectively calculated, given that the true label $y^{(n)}$ of the $n$-th training sample is $1$ for positive and $0$ for negative as follows:
\begin{align}
\label{eq:selective_inter_context}
    \widetilde \L_{\text{c}}^{\text{inter}} &= \frac{1}{N_\text{P}}\sum_{n=1}^{N}{\frac{y^{(n)}}{H(H-1)} \norm{\mathbf{C}^{(n)\text{T}}\mathbf{C}^{(n)} - \mathbf{I}_H}^2_\text{F}} \\
\label{eq:selective_intra_context}
    \widetilde \L_{\text{c}}^{\text{intra}} &= \frac{1}{H}\sum_{i=1}^{H}{\frac{1}{N_\text{P}(N_\text{P}-1)} \norm{\mathbf{Y}(\widetilde{\mathbf{C}}_i^\text{T}\widetilde{\mathbf{C}}_i - \mathbf{I}_{N})\mathbf{Y}}^2_\text{F}} \\
\label{eq:selective_inter_score}
    \widetilde \L_{\text{s}}^{\text{inter}} &= \frac{1}{N_\text{P}}\sum_{n=1}^{N}{\frac{y^{(n)}}{H(H-1)} \norm{\mathbf{E}^{(n)\text{T}}\mathbf{E}^{(n)} - \mathbf{I}_H}^2_\text{F}},
\end{align}
where ${N_\text{P}}$ denotes the number of positive samples and $\mathbf{Y}$ is the diagonal selection matrix $\text{diag}(y^{(0)}, y^{(1)}, ..., y^{(N)})$.

Now we can write the problem as minimization of the cross entropy loss with the regularization terms as follows:
\begin{equation}
\label{eq:problem}
    {\mathbf{\theta}}^*=\underset{\mathbf{\theta}}{\mathrm{argmin}} \big\{ \L_{\text{CE}}+\lambda_1 \widetilde  \L_{\text{c}}^{\text{inter}} - \lambda_2 \widetilde \L_{\text{c}}^{\text{intra}} +\lambda_3 \widetilde \L_{\text{s}}^{\text{inter}}  \big\},
\end{equation}
where each $\lambda_i$ is a hyperparameter that controls the importance of the corresponding regularization term. Note that $\L_{\text{c}}^{\text{intra}}$ has the opposite sign, since this regularization term is to be maximized while the others are to be minimized.




\subsection{Semi-supervised salience learning}
One interesting perspective of this work is that it roughly provides a semi-supervised way of learning representations of salient features from keyword utterances for the given task. In other words, without the sequence part alignment information such as phoneme labels and frame indices, the encoder finds task-relevant subsequences which have important roles for distinguishing keywords from non-keywords while only the keyword label is provided. Fig. \ref{fig:attention_example} illustrates examples of attention weights for an utterance of the ``Hey Snapdragon'' keyword. In Fig. \ref{fig:attention_example}(a) and (b), it can be seen that the single head attention has a wide range of weight distribution across time with emphasis on the keyword end part, while the attention weights from different heads of the plain, i.e., without regularization, multi-head model are distributed in different positions capturing the encoder output representations of the corresponding subsequences. However, some of them overlap with each other, indicating the context vectors from the attention heads have redundant information. With the proposed regularization, it can be seen in Fig. \ref{fig:attention_example}(c) that the attention heads pay attention to exclusive sequence parts.


\section{EXPERIMENTS}
\label{sec:exp}

\subsection{Datasets and experimental setup}
The target keyword in our experiments is ``Hey Snapdragon'' which consists of four English syllables.
In order to train the model and evaluate the performance, we collected a number of clean positive and negative samples from 325 speakers. The positive dataset has $\sim$12,000 samples from 325 speakers and divided into training, validation and test subsets at a ratio of 10:1:1. For validation and test datasets, we augmented the keyword utterances with 4 types of noises, i.e., babble, car, music, office, at signal-to-noise ratios (SNRs) of -6, 0, and 6 dB and with reverberation with a room impulse response measured in a regular meeting room, so that the total number of each of the positive validation and test samples is $\sim$15,000. For negative samples, we collected $\sim$400 hours of general English sentences and divided them at a ratio 1:1:1 for training, validation, and test. We also augmented the negative validation and test datasets with random noises to double the amount, so that the total number of each of the negative validation and test samples is $\sim$38,000 and $\sim$33,000, respectively. Note that there is no duplication and no overlap in speaker, noise sample and room impulse response between all positive and negative training, validation, and test sets.

To improve acoustic environmental robustness, we augmented 50\% of positive and negative training samples in an online manner where each sample is synthetically corrupted during data loading with randomly selected room impulse response and background noise sample from of $\sim$200 hours of noise and reverberation datasets. We assumed that all data have a fixed length and thus segmented them to 1.8 s length while guaranteeing all utterance in the training set are not clipped out in time. This assumption does not restrict on-device usability as we can apply sliding window techniques in continuous audio stream without harming the assumption. From 1.8 s input audio sequences sampled at 16 kHz, 40-dimensional Mel filter bank energies with per-channel energy normalization \cite{wang2017trainable} were computed for 30 ms frames at every 10 ms by performing short-time Fourier transform with 512-point Hamming window, and then fed into the network. 

We performed experiments with the network structure described in \ref{subsec:base_system} while varying the number of attention heads with empirically chosen $\lambda$ values in (\ref{eq:problem}). All models were trained from scratch with randomly initialized parameters for 200 epochs which is considered to be a sufficient number to reach convergence. A mini-batch was constituted with 128 shuffled positive and negative training samples with their numbers of ratio 1:3. We used Adam optimizer \cite{kingma2014adam} with a learning rate of $2\times10^{-4}$ which decays at each epoch with a factor of 0.98 while gradients with norm values above 1.0 were clipped. Since each attention head has learnable parameters in scoring function \ref{eq:scoring_function} and the number of nodes in the softmax layer changes due to concatenation of the context vectors from the attention heads, the number of parameters of 4-head model is 91 k while that of the single-head model is 78 k.

\subsection{Regularization loss variation}
\label{subsec:reg_loss}

\begin{figure}[t!]
    \begin{minipage}[b]{1.0\linewidth}
    \centering
    \subfigure[]{\includegraphics[width=0.99\linewidth]{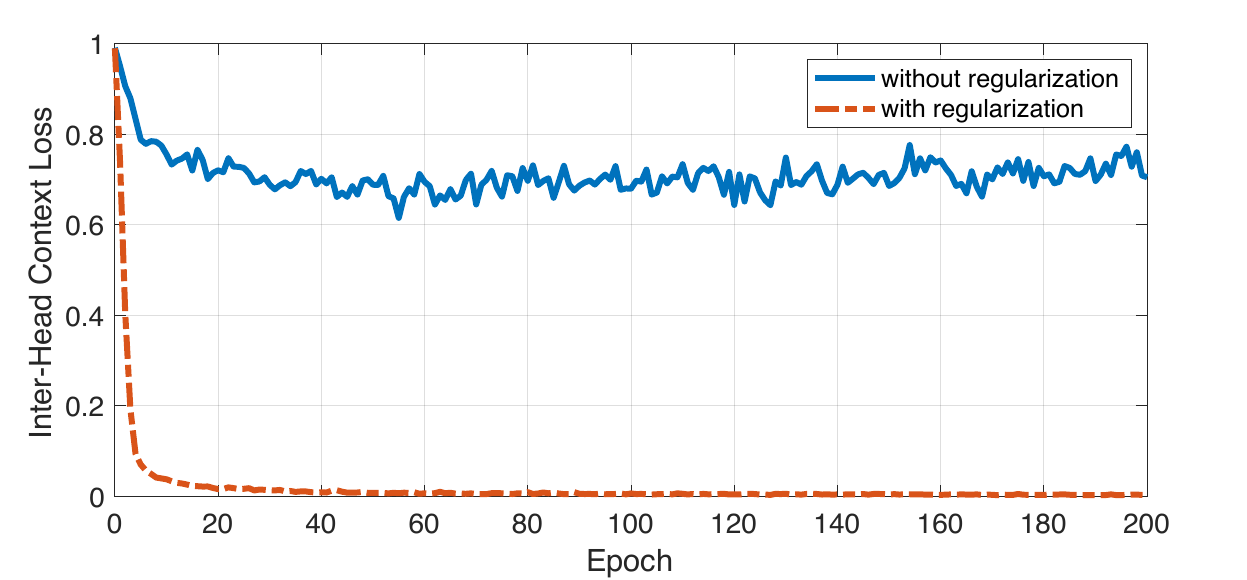}}
    \subfigure[]{\includegraphics[width=0.99\linewidth]{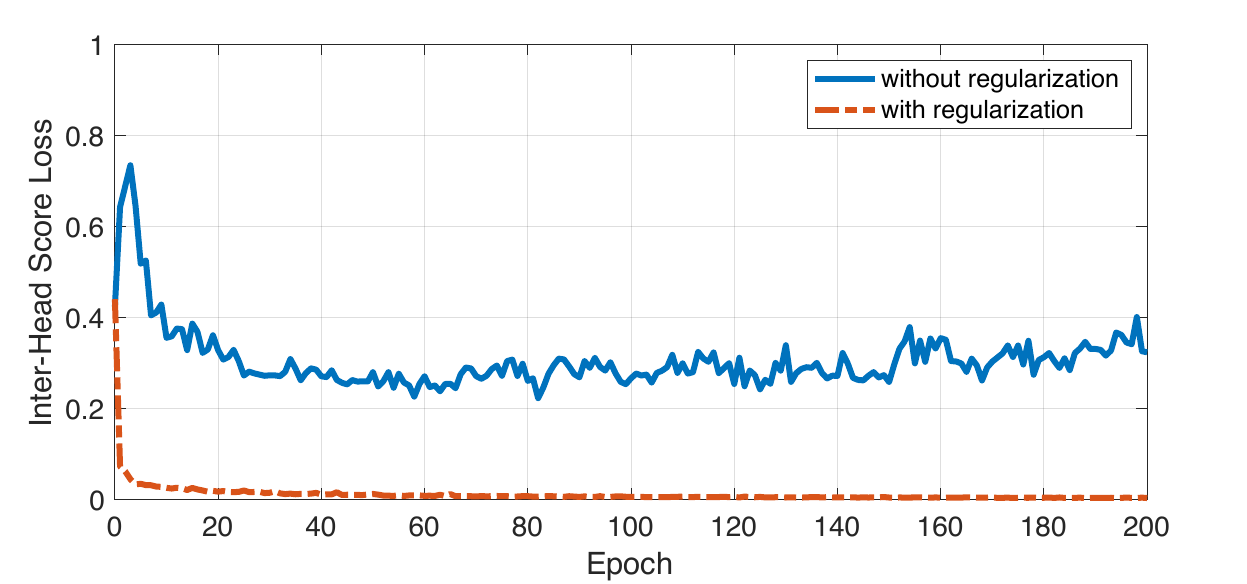}}
    \subfigure[]{\includegraphics[width=0.99\linewidth]{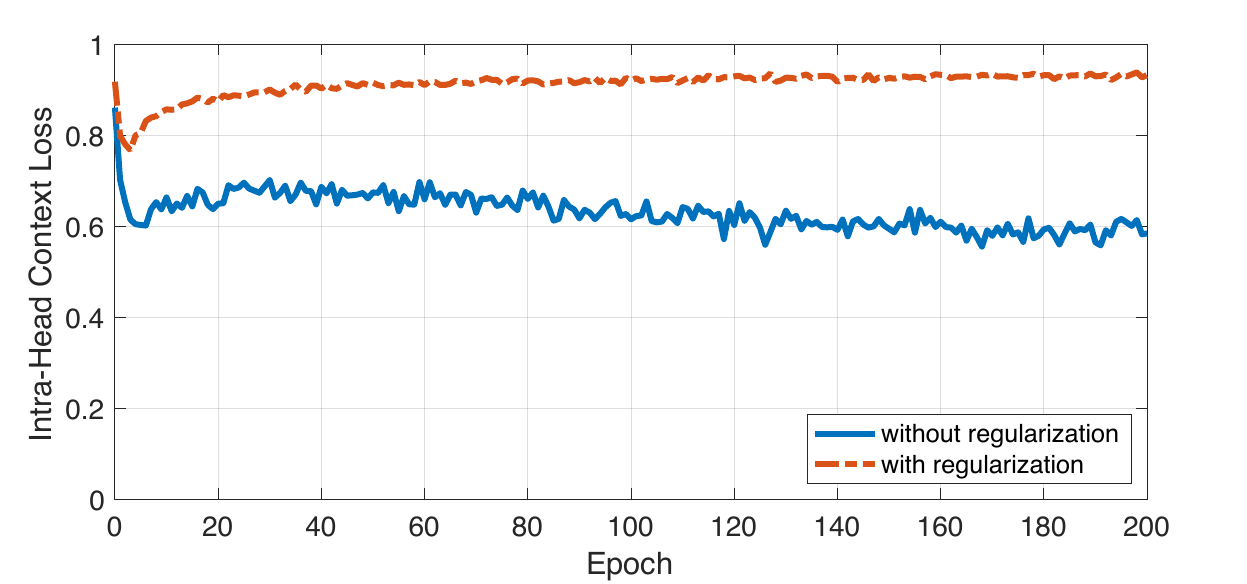}}
    \end{minipage}
    \caption{Regularization losses of (a) inter-head context orthogonality $\L_{\text{c}}^{\text{inter}}$, (b) inter-head score orthogonality $\L_{\text{s}}^{\text{inter}}$, and (c) intra-head context orthogonality $\L_{\text{c}}^{\text{intra}}$. All losses are calculated for validation sets during training.}
    \label{fig:reg_loss}
\end{figure}

Fig. \ref{fig:reg_loss} shows the regularization losses calculated from the positive validation set during training. It can be observed that $\L_{\text{c}}^{\text{inter}}$ and $\L_{\text{s}}^{\text{inter}}$ are decreasing as intended, i.e., the orthogonality between the context vectors and the score vectors between the attention heads are increasing, meaning that inter-head redundancy in time and subspace is reduced by the regularization. Meanwhile, as can be seen in Fig. \ref{fig:reg_loss}(c), $\L_{\text{c}}^{\text{intra}}$ increases which indicates that the output context vectors of each attention head from different positive samples get more similar to each other as training progresses. This is desirable for the classification stage because, generally, it is beneficial to have less variation of feature representation, i.e., context vector, in feature space for the positive samples.

\subsection{Performance with different combinations of regularizations}
\label{subsec:reg_combination}

To see how the regularization affects the keyword spotting performance, we compare the keyword spotting test results for different combinations of regularizations applied during training. False rejection rates (FRR) measured at confidence thresholds corresponding to 1 false alarm per hour (FA/hr) for corresponding models are used for the performance metric. For simplicity of comparison, we fixed the number of attention heads as 4, motivated by the keyword has 4 syllables, and the $\lambda$ value as 0.1.

\begin{table}[!t]
\centering
\caption{Validation performance of different regularization configurations measured by FRR (\%) at 1 FA/hr. $^{(*)}$ indicates that the selective regularization is not applied.}
\begin{tabular}{cccc|ccc}
\thickhline
\multicolumn{4}{c|}{Systems}   & \multicolumn{3}{c}{FRR (\%) at} \\ \hline
$H$ & $\lambda_{1}$   & $\lambda_{2}$   & $\lambda_{3}$   & 1 FA/hr  & 2 FA/hr  & 4 FA/hr  \\ \hline
1   & -               & -               & -               & 5.57    & 4.33    & 3.24    \\
4   & -               & -               & -               & 5.22    & 4.04    & 3.13    \\
4   & 0.1             & -               & -               & 4.70    & 3.79    & 3.00    \\
4   & -               & 0.1             & -               & 4.37    & 3.21    & 2.40    \\
4   & -               & -               & 0.1             & 4.58    & 3.58    & 2.75    \\
4   & 0.1             & 0.1             & -               & 4.44    & 3.46    & 2.59    \\
4   & 0.1             & -               & 0.1             & 3.97    & 2.97    & 2.27    \\
4   & -               & 0.1             & 0.1             & 4.07    & 3.26    & 2.37    \\
4   & 0.1             & 0.1             & 0.1             & \bf{3.91}    & \bf{2.88}    & \bf{2.07} \\ \hline
$^{(*)}$4 & 0.1       & 0.1             & 0.1             & 5.50    & 4.17    & 3.05    \\ \thickhline
\end{tabular}
\label{table:reg_combination}
\end{table}

From Table \ref{table:reg_combination}, we can see that all types of regularization contributes for improving the performance both individually and in combination, while using all regularization terms gives the lowest FRR. Note that using plain multi-head attention also gives some improvement over the single head attention model. At the thresholds corresponding to 1 FA/hr, the proposed regularization introduces up to 32.6\% and 25.1\% relative reduction of FRRs over the single head attention model and the plain multi-head attention model, respectively. 

\begin{figure}[t]
  \centering
  \includegraphics[width=0.99\linewidth]{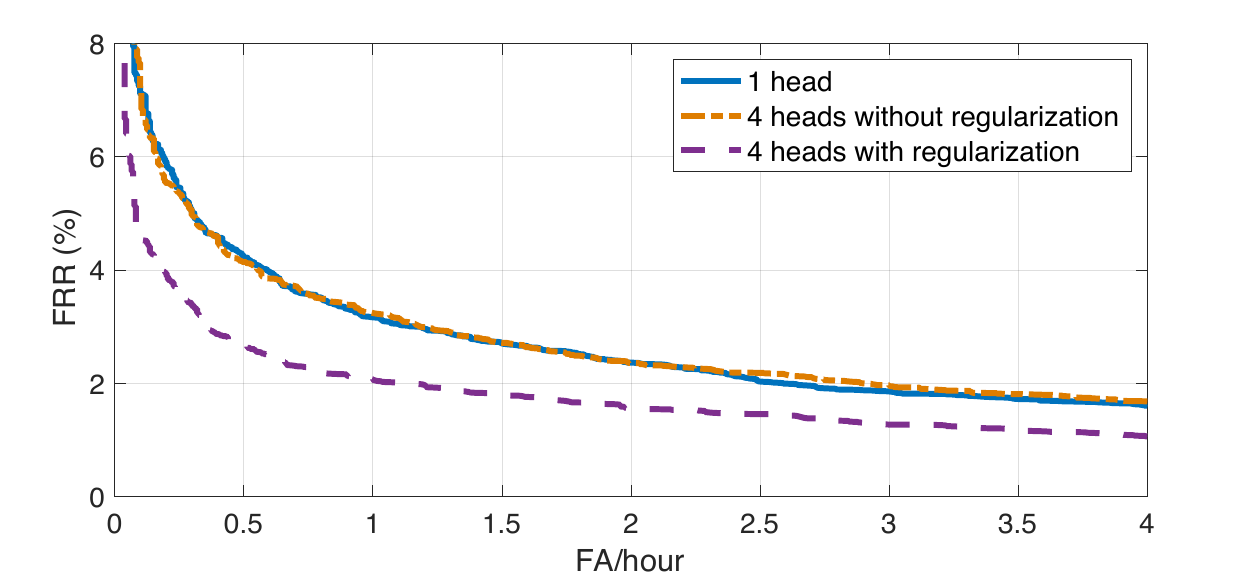}
  \vspace*{-10pt}
  \caption{Test ROC curves of models with single and 4 attention heads. The numbers in the bracket are ($\lambda_1$, $\lambda_3$, $\lambda_3$). In each configuration, the model with the lowest FRR at 1 FA/hr is chosen for comparison.} 
  \vspace*{-5pt}
  \label{fig:roc_combination}
\end{figure}

Fig. \ref{fig:roc_combination} shows that the receiver operating characteristic (ROC) curves of the single-head, the plain 4-head, and the regularized 4-head attention models for the test dataset where we set all $\lambda$'s to 0.1. It can be seen that the regularized multi-head model consistently and significantly outperforms both the single-head attention model and the plain or non-regularized multi-head attention model for all FA/hr. At 1 FA/hr, for the test dataset, FRRs are reduced by 34.4\% and 36.0\%, respectively.

\subsection{Varying $\lambda$ values}
\label{subsec:varying_lambda}

\begin{figure}[t]
  \centering
  \includegraphics[width=0.99\linewidth]{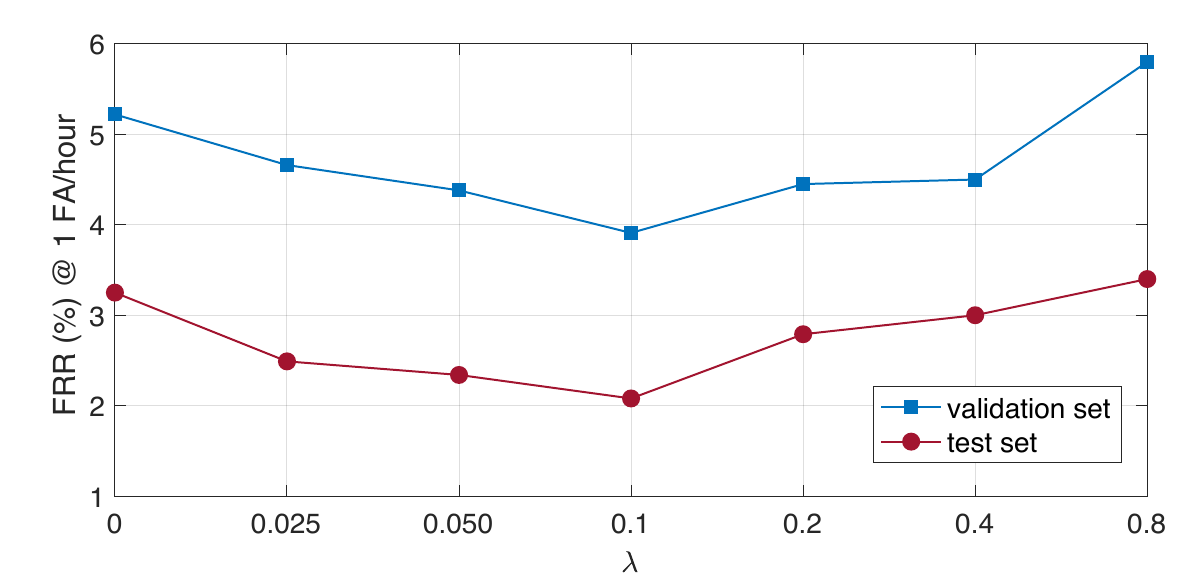}
  \vspace*{-10pt}
  \caption{FRRs at 1 FA/hr with varying $\lambda$ values while all $\lambda$'s are set to the same value.}
  \vspace*{-5pt}
  \label{fig:varying_lambda}
\end{figure}

We also show how the performance changes according to the $\lambda$ values. To see the change, we varied the $\lambda$ values from $0$ to $1.0$ while all $\lambda$ have the same value in one training instance for simplicity. The number of attention heads is fixed to $4$ as before. From Fig. \ref{fig:varying_lambda}, we can see that the best performance is achieved at $\lambda=0.1$. Although we did not investigate the different combinations of $\lambda$ values for different regularization terms, this result suggests that one can find the optimal point in the hyperparameter space of $\lambda$'s for which automated machine learning algorithms can be used.

\section{CONCLUSION}
\label{sec:conclusion}

In this paper, we have proposed a multi-head attention-based keyword spotting system trained with regularization derived from orthogonality constraints on context and score vectors of attention heads. The inter-head orthogonality regularization of context vectors and score vectors encourages the attention heads to have less redundancy to each other in positions and subspaces, while the intra-head non-orthogonality regularization of context vectors lets them have contextual consistency across samples for the given task. The proposed orthogonality constrained multi-head attention mechanism has been shown to learn exclusive representation of sequence parts both in position and in subspaces, which in turn improves the keyword spotting performance by extracting richer task-relevant information from structured data. In the experiment with the ``Hey Snapdragon" keyword, the proposed method reduced the relative false rejection rate by 34.4\% and 36.0\% at 1 FA/hr over single-head and plain multi-head attention-based models, respectively, for the test dataset. Our future works include investigation on other criteria for regularizing multi-head attention and extension of the idea to other speech tasks such as speaker verification and speech recognition.

\vfill
\pagebreak


\bibliographystyle{IEEEbib}
\bibliography{refs.bib}

\end{document}